\documentclass[conference]{IEEEtran}
\IEEEoverridecommandlockouts
\usepackage{graphicx}
\usepackage{multirow}
\usepackage{booktabs}
\usepackage[table]{xcolor}
\usepackage{cite}
\usepackage{amsmath,amssymb,amsfonts}
\usepackage{algorithmic}
\usepackage{graphicx}
\usepackage{textcomp}
\usepackage{xcolor}
\def\BibTeX{{\rm B\kern-.05em{\sc i\kern-.025em b}\kern-.08em
    T\kern-.1667em\lower.7ex\hbox{E}\kern-.125emX}}

\DeclareRobustCommand*{\IEEEauthorrefmark}[1]{\raisebox{0pt}[0pt][0pt]{\textsuperscript{\footnotesize\ensuremath{#1}}}}
\begin{document}

\title{Towards General Multimodal Visual Tracking
}

\author{
	\IEEEauthorblockN{
		Andong Lu\IEEEauthorrefmark{1}, 
		Mai Wen\IEEEauthorrefmark{2}, 
            Jinhu Wang\IEEEauthorrefmark{1},
            Yuanzhi Guo\IEEEauthorrefmark{2},
		Chenglong Li\IEEEauthorrefmark{2}*, 
		Jin Tang\IEEEauthorrefmark{1} 
		and Bin Luo\IEEEauthorrefmark{1}} 
	\IEEEauthorblockA{\IEEEauthorrefmark{1}School of Computer Science and Technology, Anhui University}
	\IEEEauthorblockA{\IEEEauthorrefmark{2}School of Artificial Intelligence, Anhui University\\  adlu\_ah@foxmail.com}
    
}


\maketitle

\begin{abstract}
Existing multimodal tracking studies focus on bi-modal scenarios such as RGB-Thermal, RGB-Event, and RGB-Language. Although promising tracking performance is achieved through leveraging complementary cues from different sources, it remains challenging in complex scenes due to the limitations of bi-modal scenarios. In this work, we introduce a general multimodal visual tracking task that fully exploits the advantages of four modalities, including RGB, thermal infrared, event, and language, for robust tracking under challenging conditions. To provide a comprehensive evaluation platform for general multimodal visual tracking, we construct QuadTrack600, a large-scale, high-quality benchmark comprising 600 video sequences (totaling 384.7K high-resolution (640×480) frame groups). In each frame group, all four modalities are spatially aligned and meticulously annotated with bounding boxes, while 21 sequence-level challenge attributes are provided for detailed performance analysis. Despite quad-modal data provides richer information, the differences in information quantity among modalities and the computational burden from four modalities are two challenging issues in fusing four modalities. To handle these issues, we propose a novel approach called QuadFusion, which incorporates an efficient Multiscale Fusion Mamba with four different scanning scales to achieve sufficient interactions of the four modalities while overcoming the exponential computational burden, for general multimodal visual tracking. Extensive experiments on the QuadTrack600 dataset and three bi-modal tracking datasets, including LasHeR, VisEvent, and TNL2K, validate the effectiveness of our QuadFusion.
\end{abstract}

\section{Introduction}
\begin{figure*}[t]
\centering
\includegraphics[width=2\columnwidth]{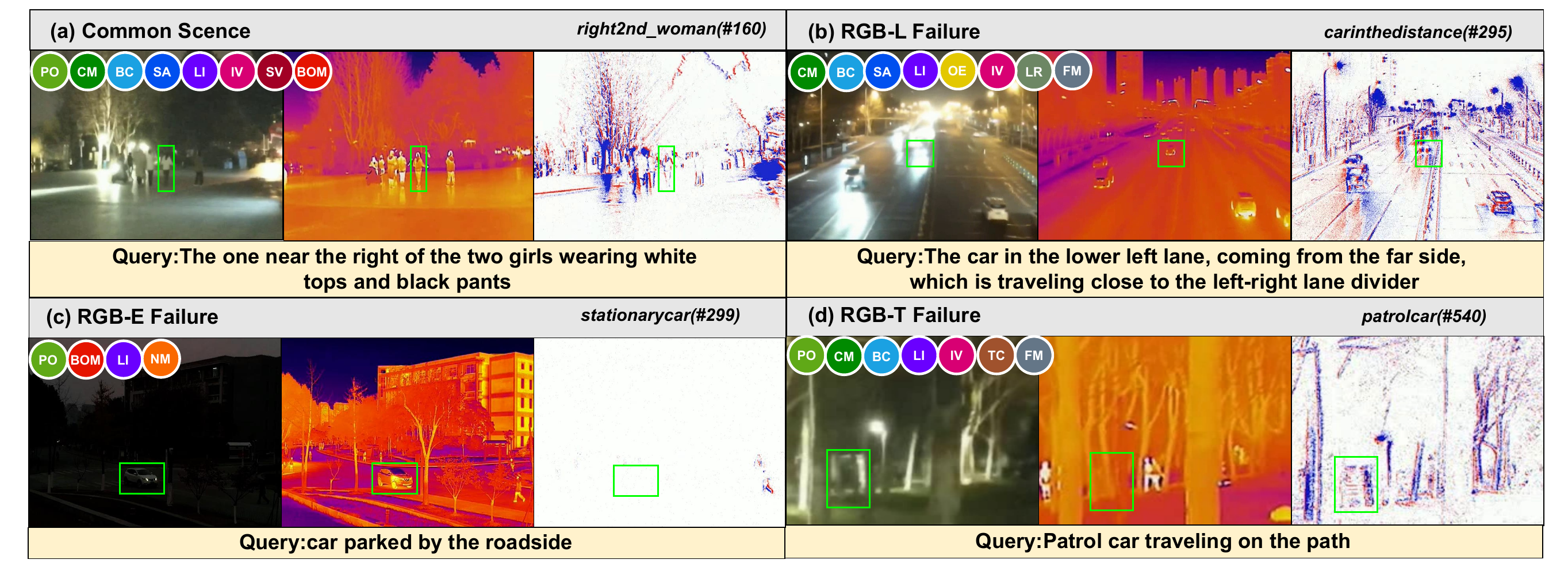}  
\caption{Some representative samples from QuadTrack600, with the challenge attributes of the sequence listed above the data, include partial occlusion (PO), camera motion (CM), background clutter (BC), similar appearance (SA), low illumination (LI), illumination variation (IV), scale variation (SV), background object motion (BOM), no motion (NM), overexposure (OE), low resolution (LR), fast motion (FM), and thermal crossover (TC). Representative samples of existing multimodal visual tracking failure scenarios are shown in (b), (c), and (d), respectively.}
\label{fig1}
\end{figure*}

Object tracking, a critical task in computer vision, focuses on identifying and following objects of interest across a video sequence. It is fundamental to numerous applications such as surveillance, autonomous vehicles, and human-computer interaction. Traditional object tracking algorithms predominantly rely on the visible spectrum, which can be severely limited under challenging conditions such as low light, glare, and haze.

To overcome these limitations, recent studies integrate additional modalities to enhance tracking performance. For instance, multimodal tracking approaches such as RGB-Thermal (RGB-T), RGB-Event (RGB-E), and RGB-Language (RGB-L) leverage complementary information to improve tracking robustness. Specifically, early RGB-T datasets like LITIV~\cite{lahmyed2019new} provide only limited sequences, while subsequent datasets~\cite{li2016gtot,li2017weighted,li2019rgb,li2021lasher} expand the scale and complexity of the benchmarks. Similarly, RGB-E tracking evolves from experiments on simulated data~\cite{hu2016dvs,liu2016combined} to substantial datasets such as FE108~\cite{zhang2021object} and VisEvent~\cite{wang2023visevent}. In the field of RGB-L tracking, pioneering works such as OTB-LANG~\cite{li2017tracking} pave the way for larger datasets like LaSOT~\cite{fan2019lasot} and TNL2K~\cite{wang2021towards}. 
Despite these advancements, bi-modal tracking approaches remain inherently limited when addressing complex real-world scenarios, where extreme lighting, occlusion, and adverse weather conditions occur simultaneously, as shown in Fig.~\ref{fig1}.
This limitation motivates the introduction of a wider range of modalities to improve tracking accuracy and robustness.

To this end, we introduce QuadTrack600, a large-scale benchmark for general multimodal visual tracking that integrates four distinct modalities: visible, thermal infrared, event data, and language. QuadTrack600 is the first dataset to incorporate four modalities, offering high diversity and complexity. In particular, QuadTrack600 comprises 600 video sequences, each containing visible, thermal infrared, and event data, as well as a language description of the target in the initial frame, totaling over 730K frame pairs. Each frame pair is spatially aligned and manually annotated with bounding boxes, ensuring high-quality annotations. This dataset is pivotal for both research and comprehensive evaluation of general multimodal visual tracking methods. Notably, despite its important role in measuring depth information, the depth modality is mainly applied to indoor scenes~\cite{zhu2023rgbd1k,zhu2024unimod1k}, while this paper focused on establishing a benchmark for open scenes.

A general multimodal visual tracking benchmark presents both challenges and opportunities beyond bi-modal methods. 
%
%
Bi-modal tracking fuses two complementary data types to boost accuracy, while integrating four distinct modalities offers a much richer, more robust source of information. However, the increased diversity comes with two critical challenges.
First, each modality delivers different types and quantities of information: RGB offers high-resolution details, thermal provides moderate yet robust cues in low light, event data delivers coarse dynamic signals, and language varies from concrete to abstract. Existing fusion techniques~\cite{cao2024bi,zhu2023visual,ma2024unifying}, often designed for two modalities, fail to adaptively handle these differences. 
Second, integrating four data streams significantly increases computational complexity, as Transformer-based fusion methods~\cite{hui2023bridging,zhu2023cross,zhang2023all} incur exponential costs from long token interactions.

To handle these issues, we propose a novel approach, called QuadFusion, for general multimodal visual tracking. QuadFusion integrates a Multiscale Fusion Mamba module that employs four distinct scanning scales: modal-level forward and backward scanning, region-level scanning, and token-level scanning.
This design not only adaptively balances the disparate information quantities across modalities but also confines inter-modal interactions to a block with linear complexity, thereby mitigating the computational burden.
Extensive experiments on QuadTrack600 benchmark, along with evaluations on existing bi-modal datasets, validate the effectiveness and generalizability of QuadFusion. Our main contributions are summarized as follows:

\begin{itemize}
\item Different from existing bi-modal tasks, we build a new general multimodal visual tracking task, which is an important one of the trends for the future development of multimodal visual tracking.

\item We introduce the first large-scale general multimodal visual tracking benchmark QuadTrack600, which provides 600 high-quality four modal data containing RGB, thermal infrared, events, and language for robust tracking. Comprehensive evaluation and analysis of mainstream bi-modal trackers suggest that QuadTrack600 is highly challenging and significant.

\item We propose a new general multimodal visual tracking approach QuadFusion, 
which efficiently and adaptively fuses four heterogeneous modalities through a Multiscale Fusion Mamba, effectively balancing information diversity while avoiding exponential computational overhead.

\item Extensive experiments on multiple multimodal benchmarks and tracking frameworks validate the effectiveness and generalization of QuadFusion. 
\end{itemize}

\section{Related Work}

\subsection{Multimodal Visual Tracking}
The field of specific multimodal tracking has developed rapidly in recent years, which mainly includes visible and thermal infrared (RGB-T) tracking, visible and event (RGB-E) tracking, and visible and language (RGB-L) tracking.
In \textbf{RGB-T tracking}, Li et al.~\cite{li2021lasher} construct the current largest RGBT tracking dataset containing 1,224 sequences, which attracts much attention to this field. For instance, Cao et al.~\cite{cao2024bi} propose a bi-directional adapter to achieve mutual prompting of inter-modal information in an adaptive manner. Tang et al.~\cite{tang2024generative} first design a diffusion-based generative fusion strategy to enhance the training effect of modality fusion. Additionally, many other works~\cite{hui2023bridging,zhu2023visual,zhang2023efficient,yang2022prompting} have significantly contributed to the development of the field.
In \textbf{RGB-E tracking}, Wang et al.~\cite{wang2023visevent} construct one of the most influential RGB-E tracking dataset, which contributes to the development of RGB-E tracking~\cite{zhu2023cross,zhang2022spiking,zhang2023frame,wang2024event}. Among them, Zhang et al.~\cite{zhang2023frame} introduce a event-guided alignment framework to perform both cross-modal and cross-frame-rate alignment, which achieving high frame-rate tracking. Wang et al.~\cite{wang2024event} propose a novel hierarchical cross-modality knowledge distillation strategy for event-based tracking, transferring RGB-Event knowledge to an unimodal event-based tracker. 
In \textbf{RGB-L tracking}, TNL2K~\cite{wang2021towards} is an important RGB-L tracking dataset containing 2000 sequences with target language descriptions, which contributes to many excellent works~\cite{zhou2023joint,li2023citetracker,zheng2023towards,zhang2023all}. Recently, Ma et al.~\cite{ma2024unifying} propose a modality-unified feature framework to joint model visual and language modalities. Shao et al.~\cite{shao2024context} propose a novel QueryNLT to integrate various modality references for target modeling and matching, which enhances the overall understanding and discrimination of targets. 
Although these specific multimodal visual tracking studies succeed significantly, the bi-modal data are still limited when confront with highly complex scenarios. Moreover, different bi-modal tracking studies are hard to compare each other due to different tracking benchmarks. In this work, we first presents more general multimodal visual tracking benchmarks. 

\subsection{Mamba-based Models in Visual Tasks}
Since its introduction for linear-time sequence modeling in NLP, Mamba~\cite{mamba} has rapidly extended to various computer vision tasks. Vmamba~\cite{zhu2024vision} adopts a four-way scanning algorithm tailored for image features, outperforming Swin-Transformer~\cite{liu2021swin} in detection, segmentation, and tracking. VM-UNet~\cite{ruan2024vm} excels in medical segmentation by integrating Mamba blocks into the UNet framework. Video-Mamba~\cite{li2024videomamba} expands 2D scanning to bidirectional 3D scans, enhancing video understanding. Recent advancements include PlainMamba~\cite{yang2024plainmamba} for 2D continuous scanning, LocalMamba~\cite{huang2024localmamba} for dynamic layer-specific scanning, and RSMamba~\cite{chen2024rsmamba} for remote sensing tasks. Mamba-based methods have also been applied to infrared small target detection \cite{chen2024mim} and spatio-temporal relationship learning in bi-temporal data \cite{chen2024changemamba}.
In this work, we propose a new Multiscale Fusion Mamba tailored for general multimodal visual tracking, which builds a progressive scanning granularity for inter-modal fusion.

\section{QuadTrack600 Benchmark Dataset}
\begin{table*}[h]
    \centering
    \caption{Comparison of QuadTrack600 with existing bi-modal tracking benchmarks.}
    \setlength{\tabcolsep}{2mm}
    
    \renewcommand\arraystretch{1.4}{
    \resizebox{1\textwidth}{!}{
    \begin{tabular}{c|c|c|cccc|c|c|c|c|c|c|c}
    \toprule
    \multirow{2}{*}{\textbf{Benchmark}} & \multirow{2}{*}{\textbf{Pub. Info}} & \multirow{2}{*}{\textbf{Task-oriented}} & \multicolumn{4}{c|}{\textbf{Modalities}}  & \multirow{2}{*}{\textbf{Sequence num}} & \multirow{2}{*}{\textbf{Average frames}} & \multirow{2}{*}{\textbf{Total frames}}
    & \multirow{2}{*}{\textbf{Class}} & \multirow{2}{*}{\textbf{Attributes}} & \multirow{2}{*}{\textbf{Resolution}}
    & \multirow{2}{*}{\textbf{TrainingSet}} \\
    & & & \textbf{RGB} & \textbf{TIR} & \textbf{Event} & \textbf{Language} & & & & & &  \\ 
    \hline\hline
    TrackingNet\cite{muller2018trackingnet} & ECCV2018 & \multirow{2}{*}{RGB} & \checkmark & - & - & - & 30.6K & 501 & 14M & 21 & 15 & 591$\times$1013 &  \checkmark \\
    GOT-10K\cite{fan2019lasot} & TPAMI2019 &  & \checkmark & - & - & - & 10K & 150 & 1.5M & 563 & 6 & 929$\times$1638 &  \checkmark \\
    \hline
    GTOT\cite{li2016gtot} & TIP2016 & \multirow{5}{*}{RGB-T} & \checkmark & \checkmark & - & - & 50 & 157 & 7.8K & 9 & 7 & 384$\times$288 & -  \\
    RGBT210\cite{li2017weighted}& ACM MM2017 & & \checkmark & \checkmark & - & - &  210 & 498 & 104.7K & 22 & 12 & 630$\times$460 & -  \\
    RGBT234\cite{li2019rgb} & PR2019 & & \checkmark & \checkmark & - & - & 234 & 498 & 116.7K & 22 & 12 & 630$\times$460 & - \\
    LasHeR\cite{li2021lasher}& TIP2021 & & \checkmark & \checkmark & - & - & 1224 & 600 & 734.8K & 32 & 19 & 630$\times$480 & \checkmark  \\
    VTUAV\cite{Zhang_CVPR22_VTUAV}& CVPR2022 & & \checkmark & \checkmark & - & - & 500 & 3329 & 1.7M & - & - & 1920$\times$1080 & \checkmark \\
    \hline
    LaSOT\cite{fan2019lasot} & CVPR2019 & \multirow{2}{*}{RGB-L} & \checkmark & - & - & \checkmark & 1400 & 2506 & 3.5M & - & 14 & - & \checkmark \\
    TNL2K\cite{wang2021towards} & CVPR2021 & & \checkmark & - & - & \checkmark & 2000 & 622 & 1.2M & - & 17 & - & \checkmark \\
    \hline
    FE108\cite{zhang2021object} & ICCV2021 & \multirow{3}{*}{RGB-E} & \checkmark & - & \checkmark & - & 108 & 1932 & 208.6K & 21 & 4 & 346$\times$260 & \checkmark \\
    COESOT\cite{tang2022revisiting} & arXiv2022 & & \checkmark & - & \checkmark & - & 1354 & 354 & 478.7K & 90 & 17 & 346$\times$260 & \checkmark \\
    VisEvent\cite{wang2023visevent} & TCyber2023 & & \checkmark & - & \checkmark & - & 820 & 452 & 371.1K & - & 17 & 346$\times$260 & \checkmark \\
    \hline
    \cellcolor{lightgray}QuadTrack600 & \cellcolor{lightgray}- & \cellcolor{lightgray}RGB-T-L-E & \cellcolor{lightgray}\checkmark & \cellcolor{lightgray}\checkmark & \cellcolor{lightgray}\checkmark & \cellcolor{lightgray}\checkmark & \cellcolor{lightgray}600 & \cellcolor{lightgray}581 & \cellcolor{lightgray}348.7K & \cellcolor{lightgray}40 & \cellcolor{lightgray}21 & \cellcolor{lightgray}640$\times$480 & \cellcolor{lightgray}\checkmark \\
    \bottomrule
    \end{tabular}}}
    
\label{data_overall_compare}
\end{table*}

To resolve the problem that existing studies focus on bi-modal visual tracking while lacking the exploration of general multimodal visual tracking, we construct a towards general multimodal visual tracking dataset, named QuadTrack600. 
Since this dataset is oriented towards general multimodal visual tracking in open scenarios, the wider imaging range of RGB, TIR and Event modalities are selected, instead of the depth or sonar modality, which is mainly used indoors or underwater.
As a result, QuadTrack600 comprises 600 pairs of quad-modal data, i.e., RGB, TIR, Event, and Language, with a total frame count reaching 348K.
The major properties of QuadTrack600 over existing bi-modal tracking datasets are shown in Tab.~\ref{data_overall_compare}. The details are analysed below.

\subsection{Dataset Construction}

\begin{figure}[ht]
\centering
\includegraphics[width=1\columnwidth]{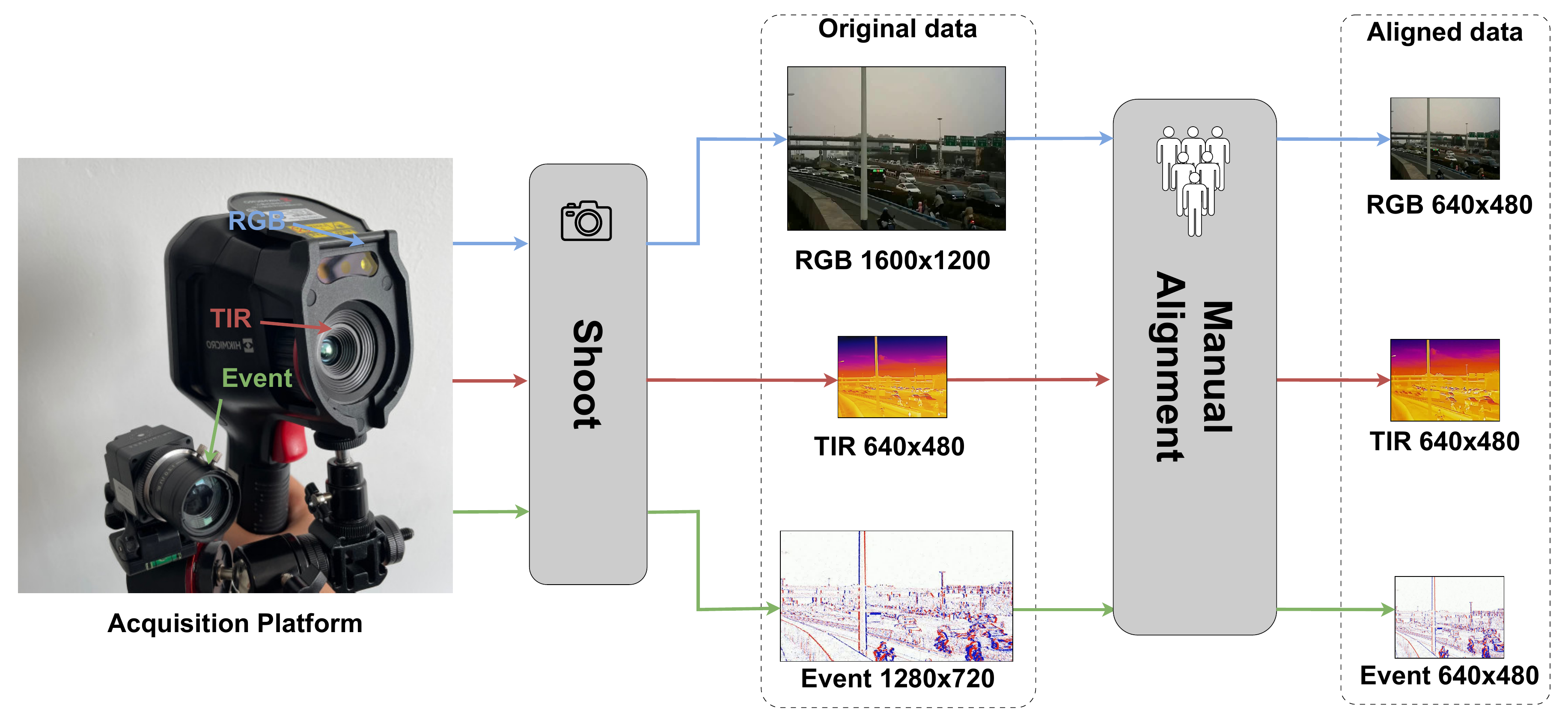} 
\caption{Workflow of data collection and data alignment. }
\label{fig2}
\end{figure}

\noindent\textbf{Data collection and alignment.} To simultaneously collect video sequences from all three modalities, we construct a handheld imaging system as shown in the acquisition platform of Fig.~\ref{fig2}. The system consists of two parts: a Hikvision binocular thermal camera for acquiring paired visible and thermal infrared video sequences, and a DVS (Dynamic Vision Sensor) for acquiring event streaming data. By manually adjusting the imaging optical axis of the device, a common view field is available for all three modalities.

Due to differences in imaging hardware among sensors, we must register multimodal video sequences both temporally and spatially. For temporal registration, since the visible and thermal infrared sequences are pre-calibrated by the imaging hardware, we can synchronize the three modalities by adjusting the event modality data. Specifically, we set a fixed time window for the event stream data, based on the recording frame rate (25 Hz) of the binocular thermal camera, and map event points to a common plane for alignment with the other modalities. By manually processing each video sequence individually, we achieve accurate temporal registration among the three modalities. For spatial registration, we convert the temporally registered three-modality video into image frames and select the thermal infrared image with the lowest resolution (640 $\times$ 480) as the registration target. We then use professional image editing software to crop and scale the visible and event images to ensure precise alignment with the thermal infrared images. As shown in Fig.~\ref{fig2}, we can finally obtain spatio-temporally registered images from the three modalities.

\begin{figure*}[t]
\centering
\includegraphics[width=1\textwidth]{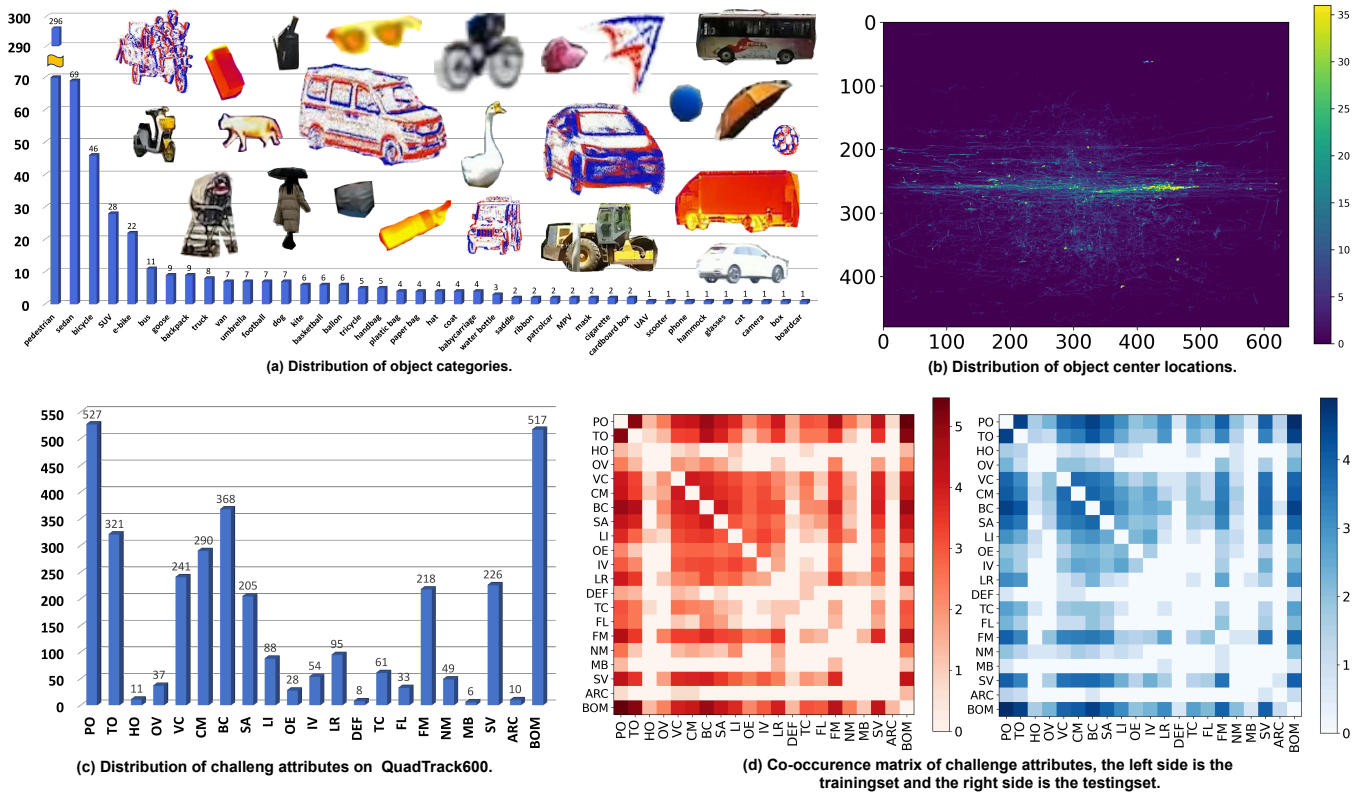} 
\caption{Some analysis and statistics on the QuadTrack600.}
\label{fig3}
\end{figure*}

\noindent\textbf{Data Annotation.}
In this dataset, we provide two complementary annotation approaches to achieve a more precise representation of the target. The first approach is the traditional bounding box (BBox) annotation, which clearly indicates the size and location of the target but can sometimes lead to ambiguity in certain scenarios. The second approach involves using language to describe the spatial location, relative positions to other objects, target attributes, categories, and other high-level semantic information.
For the BBox annotation, we utilize the ViTBAT tool to label the bounding boxes of objects frame-by-frame throughout the sequence. Note that, we adopt left corner point (x1, y1), width w and height h of the target’s bounding box are used as the ground truth, i.e., [x1, y1, w, h]. 
For the language annotation, we annotate each sequence by providing one descriptive sentence in English for the target and scene in the first frame.

\subsection{Dataset Statistics}

In this section, we statistically analyze our dataset in the following aspects, including object categories, distribution of object locations, and distribution of challenge sequences.
In particular, QuadTrack600 comprises a total of 41 target categories, in which the number of sequences for each category is depicted in Fig.~\ref{fig3} (a). It can be seen that the largest number of sequences belongs to the pedestrian category, totaling 296, followed by sequences in the transportation category (including cars, bicycles, SUVs, buses, etc.), totaling 200. We also collect sequences of animal classes (including swans, cats, and dogs), totaling 17. Moreover, we collect some other categories of targets, such as two sequences belonging to the cigarette category, two  sequences of other categories, such as two in the cigarette category, two in the mask category, and one in the glasses category, among others. The number of these sequences is relatively small compared to those in the pedestrian and transportation categories, as they are not common tracking targets.
%
Next, we plot the distribution of target locations across the entire dataset, as depicted in Fig.~\ref{fig3} (b). It can be observed that the response distribution map appears relatively scattered, suggesting that the position of the target within the videos varies significantly, and the target performs intricate movements within the scene. Such scenarios tends to increase the difficulty of the tracking task. Thus, this demonstrates the significant challenge and complexity of our proposed QuadTrack600 dataset.

In addition, we show the distribution of sequence number of challenge attributes in QuadTrack600 as shown in Fig.~\ref{fig3} (c). It can be seen that sequences with PO and BOM attributes are the most numerous, accounting for 527 and 517 sequences, respectively. This is followed by sequences with TO and BC attributes, accounting for 321 and 368 sequences, respectively. Relatively few sequences have HO, DEF, MB, and ARC attributes, accounting for 11, 8, 6, and 10 sequences, respectively. 
Lastly, we also draw the frequency distributions of co-occurring challenge attributes, separately for the training and test datasets, as shown in Fig.~\ref{fig3} (d). First, it can be observed that most sequences often contain multiple challenge attributes, which indicates the challenge of each sequence. Subsequently, it can be observed that the distribution of the training and test sets is generally consistent, indicating that the challenges of the training and test sets are highly matched.

\section{Methodology}
\begin{figure*}
\centering
\includegraphics[width=1\textwidth]{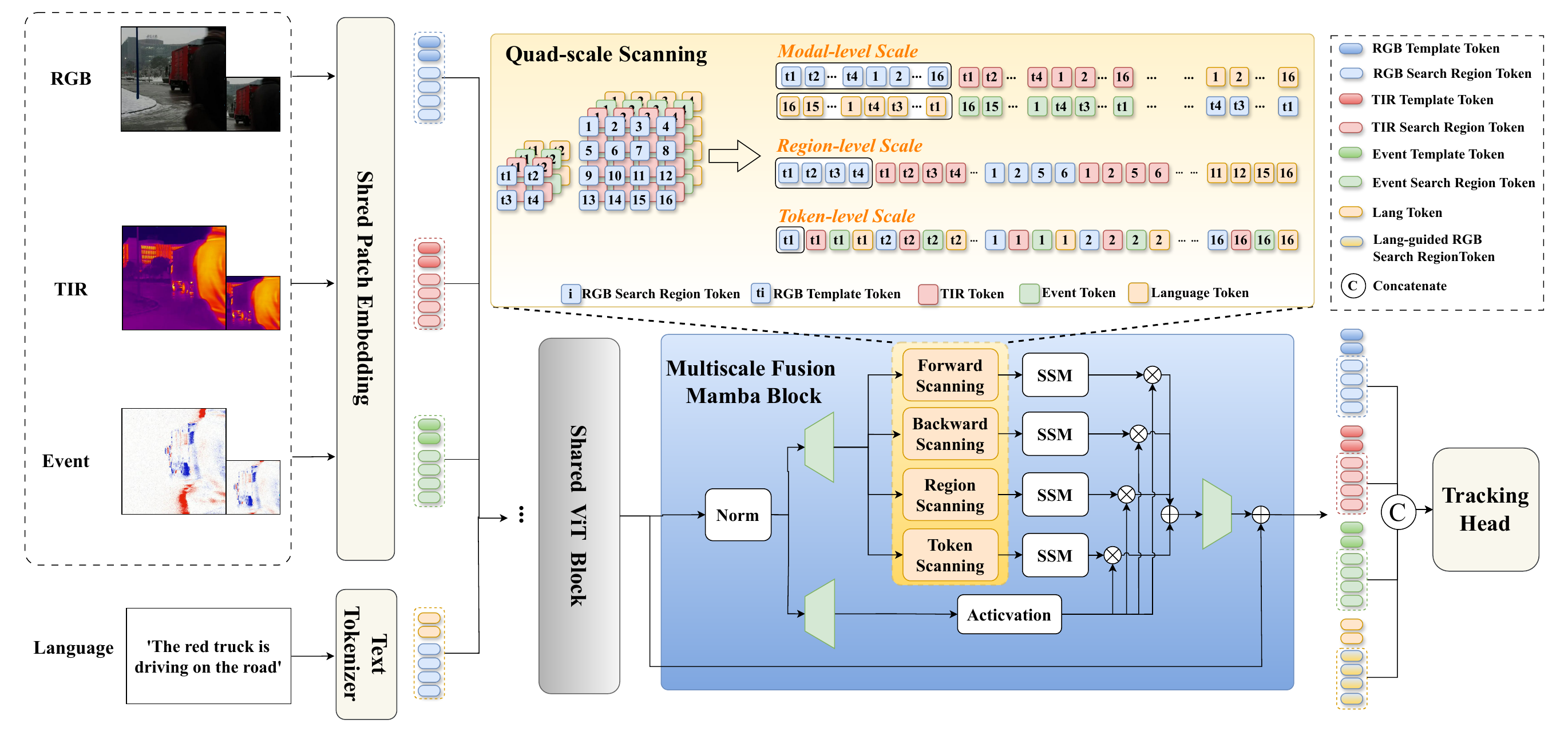}
\caption{The overall architecture of QuadFusion. First, three visual modalities and language input are embedded as tokens and processed through Transformer blocks for joint feature extraction and relationship modeling between the search and template images. In the proposed Multiscale Fusion Mamba (MFM) block, tokens from all four modalities are concatenated along the token dimension, enabling efficient multi-scale interactions across four scanning levels. Finally, the fused search-region tokens are passed to the tracking head to generate the final tracking prediction.
}
\label{pipeline}
\end{figure*}

\subsection{Overview}
The overall architecture of our proposed method is illustrated in Fig.~\ref{pipeline}, which consists of comprising two types of embedding layers, a ViT backbone, a Multiscale Fusion Mamba (MFM) block, and a tracking head. First, the search and template images from RGB, thermal, and event modalities are processed through a shared patch embedding layer, while the language modality is tokenized using a text tokenizer.
These embeddings are then combined with learnable positional embeddings to retain spatial information, represented as: $\{X_r^0,X_t^0,X_e^0\}\in \mathbb{R}^{N_x \times C}$, $\{Z_r^0,Z_t^0,Z_e^0\}\in \mathbb{R}^{N_z \times C}$, and $L^0 \in \mathbb{R}^{1 \times C}$. Here, $X$, $Z$, and $L$ represent search, template and language tokens, respectively, while $N_x$ and $N_z$ correspond to the number of search region tokens and template tokens. The subscripts $r, t, e$ correspond to RGB, thermal, and event modalities.
Next, tokens from each visual modality are concatenated along the search and template dimensions, defined as $\{H_r^0 = [Z_r^0;X_r^0], H_t^0 = [Z_t^0;X_t^0], H_e^0 = [Z_e^0;X_e^0]\} \in \mathbb{R}^{(N_z+N_x) \times C}$, where $H$ denotes the combined token embeddings for each modality. For language modality, the token embedding is concatenated only with the RGB search region tokens, since language descriptions are based on RGB content. This concatenation is expressed as $H_l^0 = [Z_l^0; X_r^0] \in \mathbb{R}^{(N_z+N_x) \times C}$, where $Z_l^0 = \text{repeat}(L^0, N_z)$, with $\text{repeat}(\cdot, N_z)$ replicates $L^0$ $N_z$ times along the feature dimension.
These token embeddings are then processed through shared ViT blocks for feature extraction and intra-modal relationship modeling between search and template frames. 
Inter-modal interactions occur within the proposed MFM block, ensuring efficient fusion across four modalities.
Finally, after passing through the ViT backbone and MFM block, the search frame features from all modalities are merged and fed into the tracking head to predict the final tracking result.

\subsection{Multiscale Fusion Mamba}
Effective inter-modal interaction is crucial for multimodal visual tracking, yet existing bi-modal methods struggle with four-modality fusion due to two key challenges.
First, achieving sufficient interaction between the four modalities is crucial for information collaboration, but the length of the token sequence (1280 in this study) will cause a large computational burden, limiting the applicability of existing Transformer-based bi-modal fusion methods~\cite{hui2023bridging,zhu2023cross,zhang2023all}.
Second, current methods~\cite{liu2024rgbt,cao2024bi,zhu2023visual,zhang2023all} ignore the differences in information quantity between different modalities, thus limiting the fine-grained fusion between different modalities.

To overcome these challenges, we design the Multiscale Fusion Mamba (MFM), a computationally efficient approach that enables full interaction among all modal tokens with linear complexity. MFM employs four distinct scanning scales to facilitate fine-grained inter-modal interactions, effectively handling the varying information quantities across the four modalities.
In particular, MFM first employs modality-level forward and backward scanning to establish global inter-modal interactions. Subsequently, it introduces region-level mixed-modal scanning based on template size to capture local dependencies. Finally, token-level mixed-modal scanning is used to achieve even more fine-grained inter-modal information exchange.
Given the output features $\{H_r^i,H_t^i,H_e^i,H_l^i\}$ from the $i$-th ViT block, the scanning process in Fig.~\ref{pipeline} is formulated as follows:
\begin{small}

\begin{equation}
\begin{aligned}
H_{forward}^i = \bigcup_{m}^{M}\bigcup_{n}^{N}H^i_m(n), 
\end{aligned}
\end{equation}

\begin{equation}
\begin{aligned}
\begin{split}
H_{backward}^i = \bigcup_{m}^{\hat{M}}\bigcup_{n}^{\hat{N}}H^i_m(n),
\end{split}
\end{aligned}
\end{equation}

\begin{equation}
 \begin{gathered}
    H_{region}^i = \left[\bigcup_{m}^{M}\bigcup_{n}^{N_z}H^i_m(n); P_{0,1}; P_{2,3}\right],\\
    P_{0,1} = \bigcup_{r}^{R}\bigcup_{m}^{M}\bigcup_{k=0}^{\sqrt{N_z}-1}\bigcup_{n=0}^{\sqrt{N_z}-1}H^i_m((2k+r+1)\sqrt{N_z}+n)\\
     P_{2,3} = \bigcup_{r}^{R'}\bigcup_{m}^{M}\bigcup_{k=0}^{\sqrt{N_z}-1}\bigcup_{n=0}^{\sqrt{N_z}-1}H^i_m((2k+r+15)\sqrt{N_z}+n)
 \end{gathered}
\end{equation}

\begin{equation}
\begin{aligned}
H_{token}^i = \bigcup_{n}^{N}\bigcup_{m}^{M}H^i_m(n).
\end{aligned}
\end{equation}
\end{small}

Here $N$=$N_z+N_x$ denotes the total number of tokens per modality, $R \in \{0,1\}$, $R'\in\{2,3\}$, and $M \in \{RGB, TIR, Event, Language\}$ represents the respective modalities. 
Finally, we merge the interaction features from all four scanning paths using a gated fusion mechanism~\cite{mamba}, producing multiscale fusion features that effectively balance inter-modal collaboration while maintaining computational efficiency.

\subsection{Implementation Details}

We adopt OSTrack~\cite{ye2022joint} as our base tracker, leveraging a ViT backbone for feature extraction. The total loss function of our approach is formulated as:
\begin{small}
\begin{equation}
 \begin{split}
    &\mathcal{L} = \mathcal{L}_{cls} + \lambda_{iou}\mathcal{L}_{iou} + \lambda_{\mathcal{L}_{1}}\mathcal{L}_{1},
\end{split} 
\end{equation}
\end{small}
where $\mathcal{L}_{cls}$ represents the weighted focal loss~\cite{law2018cornernet} for classification, while $\mathcal{L}_{iou}$ (generalized IoU loss\cite{rezatofighi2019generalized}) and $\mathcal{L}_{1}$ are used for bounding box regression. The trade-off parameters $\lambda_{iou}$ and $\lambda_{\mathcal{L}_{1}}$ follow OSTrack~\cite{ye2022joint}, set to 2.0 and 5.0, respectively. 
Our QuadFusion is implemented in PyTorch and trained on an NVIDIA A100 GPU. The model is trained for 15 epochs on the QuadTrack600 training set with a batch size of 24. The learning rate is initialized at 1e-5 and decays by a factor of 10 after 6 epochs. We use the AdamW optimizer with a weight decay of 1e-4.

\section{Experiment}
\subsection{Evaluation on QuadTrack600}
\noindent\textbf{Overall evaluation}. We evaluate QuadTrack600 using state-of-the-art trackers across different tracking domains to evaluate the challenges introduced by our benchmark.
First, we train and evaluate these trackers in a bi-modal setting using the corresponding bi-modal data from QuadTrack600, demonstrating the difficulty posed by the tracking scenarios of QuadTrack600 for existing bi-modal trackers. 
Next, we extend these trackers to quad-modal inputs by reusing their bi-modal interaction modules for four modalities, exposing their limitations in handling general multimodal visual tracking.
As shown in Tab.~\ref{table3}, performance varies across different modality combinations, yet all trackers benefit from quad-modal inputs, highlighting the advantage of integrating complementary information across multiple modalities. 
Notably, the varying degrees of improvement further reflect the adaptability of different bi-modal interaction modules to other modality combinations.
Finally, QuadFusion outperforms all trackers, achieving the best results and validating its effectiveness, emphasizing the necessity of specialized multimodal fusion designs.
\begin{table}
\centering
\caption{Performance comparison of existing state-of-the-art trackers on QuadTrack600. Each tracker is trained for optimal performance.}
\resizebox{0.5\textwidth}{!}{
\renewcommand\arraystretch{1}{
\begin{tabular}{c | c | c | c | c | c }
    \toprule
    \textbf{Method} & \textbf{Type} &  \textbf{Pub. Info.} &  \textbf{PR} & 
    \textbf{SR} \\
    \hline\hline
    OSTrack~\cite{ostrack} & RGB &  ECCV-2022 & 56.1 & 38.2 & \multicolumn{1}{|c}{\multirow{8}{*}{\rotatebox{270}{\textbf{\scriptsize Specific-modal metrics}}}}\\
     ARTrack~\cite{wei2023autoregressive} & RGB & CVPR-2023 & 54.9 & 36.0 & \multicolumn{1}{|c}{}\\
     TBSI~\cite{hui2023bridging} & RGB-T & CVPR-2023 & 59.2 & 40.0 & \multicolumn{1}{|c}{}\\
     BAT~\cite{cao2024bi} & RGB-T & AAAI-2024 & 61.5 & 42.9 & \multicolumn{1}{|c}{}\\
     ViPT~\cite{zhu2023visual} & RGB-E & CVPR-2023 & 49.9 & 33.9 & \multicolumn{1}{|c}{} \\ 
     COHA~\cite{zhu2023cross} & RGB-E & ICCV-2023 & 49.3 & 30.5 & \multicolumn{1}{|c}{}\\
     All-in-One~\cite{zhang2023all} & RGB-L & ACM MM-2023 & 44.0 & 28.4 & \multicolumn{1}{|c}{}\\
     UVLTrack~\cite{ma2024unifying} & RGB-L & AAAI-2024 & 48.8 & 33.2 & \multicolumn{1}{|c}{}\\
    \hline 
     OSTrack~\cite{ostrack} & \multirow{9}{*}{RGB-T-E-L} &  ECCV-2022 & 57.0 & 38.7 &  \multicolumn{1}{|c}{\multirow{9}{*}{\rotatebox{270}{\textbf{\scriptsize Generic-modal metrics}}}}\\
     ARTrack~\cite{wei2023autoregressive}& & CVPR-2023 & 56.4 & 38.5 & \multicolumn{1}{|c}{}\\
     TBSI~\cite{hui2023bridging} & & CVPR-2023 & 61.3 & 42.4 & \multicolumn{1}{|c}{} \\
     BAT~\cite{cao2024bi} & & AAAI-2024 & 62.6 & 43.0 & \multicolumn{1}{|c}{}\\
     ViPT~\cite{zhu2023visual} &  & CVPR-2023 & 51.5 & 35.7 & \multicolumn{1}{|c}{}\\ 
     COHA~\cite{zhu2023cross} & & ICCV-2023 & 50.9 & 35.0& \multicolumn{1}{|c}{} \\
     All-in-One~\cite{zhang2023all} & & ACM MM-2023 & 46.3 & 29.8 & \multicolumn{1}{|c}{}\\
     UVLTrack~\cite{ma2024unifying} & & AAAI-2024 & 49.3 & 34.1 & \multicolumn{1}{|c}{}\\
    \cellcolor{lightgray}QuadFusion(Ours) &  \cellcolor{lightgray}&  \cellcolor{lightgray}- & \cellcolor{lightgray}64.1 & \cellcolor{lightgray}44.4 & \multicolumn{1}{|c}{}\\
    \bottomrule
 \end{tabular}}}

\label{table3}
\end{table}

\begin{figure}[ht]
\centering
\includegraphics[width=1\columnwidth]{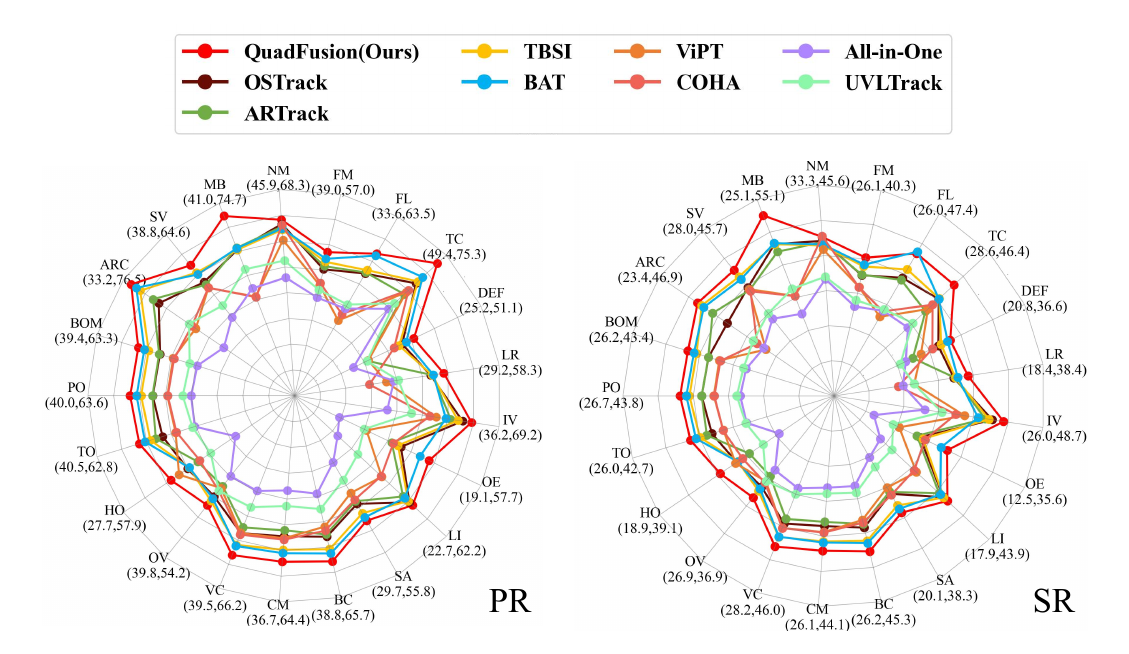}
\caption{Performance comparison of QuadFusion against advanced trackers under different challenging attributes of QuadTrack600.}
\label{atrribute_compare}
\end{figure}

\noindent\textbf{Attribute-based evaluation}. We further compare QuadFusion with leading trackers, including OSTrack\cite{ye2022joint}, ARTrack\cite{wei2023autoregressive}, TBSI\cite{hui2023bridging}, BAT\cite{cao2024bi}, ViPT\cite{zhu2023visual}, COHA\cite{zhu2023cross}, All-in-One\cite{zhang2023all}, UVLTrack\cite{ma2024unifying} on different challenge subsets. 
Fig.~\ref{atrribute_compare} presents the attribute-based evaluation, where each corner of the radar plot represents the highest and lowest performance under a specific challenge attribute. The results demonstrate that QuadFusion consistently outperforms other methods across most challenging subsets, showcasing its robustness in complex tracking scenarios.

\noindent\textbf{Evaluation on bi-modal tracking tasks}. Besides general multimodal tracking, QuadTrack600 is also suitable for specific bimodal tracking tasks. We partition QuadTrack600 into three bi-modal subsets (RGB-T, RGB-E, RGB-L) and evaluate QuadFusion alongside state-of-the-art bi-modal trackers on these subsets, as well as on three public datasets: LasHeR, VisEvent, and TNL2K. Detailed results are provided in the \textbf{Supplementary Material}.

\subsection{Ablation Studies}
\noindent\textbf{Verification of modal complementarity}.
To validate the modal complementarity within our QuadTrack600 dataset, we conduct a series of experiments using QuadFusion, with results presented in Tab.~\ref{modal_complementarity}.
\begin{table}
    \centering
    \caption{Comparison of different modal combinations.}
    \setlength{\tabcolsep}{8mm}
    \renewcommand\arraystretch{1.2}{
    \resizebox{0.5\textwidth}{!}{
    \begin{tabular}{c|c|c|c}
    \toprule
    \textbf{Input Type}& \textbf{Modalities} & {\textbf{PR}} & \textbf{SR}    \\
    \hline\hline
     \multirow{1}{*}{Unimodal} & RGB &56.9 & 39.7  \\
     \hline
      \multirow{3}{*}{Bi-modal}& RGB+T&  62.6 & 43.1 \\
    & RGB+E &  61.7 &  42.4 \\
     & RGB+L& 59.3 & 40.8 \\
     \hline
     \multirow{3}{*}{Tri-modal}& RGB+T+E & 63.7 & 43.9 \\
     & RGB+T+L& 62.9 & 43.2 \\
      & RGB+E+L& 62.0 &  42.7\\
     \hline
    \multirow{1}{*}{Quad-modal}& RGB+T+E+L & 64.1 & 44.4 \\
    \bottomrule
    \end{tabular}}}
    
\label{modal_complementarity}
\end{table}
The performance is lowest when using only the RGB modality, with PR/SR of 56.9$\%$/39.7$\%$. As we incorporate additional modalities, the tracker consistently improves: Bi-modal inputs yield an average increase of 4.3$\%$/2.4$\%$, Tri-modal inputs further enhance performance by 6.0$\%$/3.6$\%$, and Quad-modal inputs achieve the highest gains of 7.2$\%$/4.7$\%$ compared to Unimodal input.
These results demonstrate that while additional modalities complement RGB, existing bi-modal and tri-modal combinations remain insufficient for challenging scenarios, highlighting the necessity of general multimodal tracking.

\noindent\textbf{Verification of model components}. To investigate the effects of multiscale scanning strategies within the proposed Multiscale Fusion Mamba (MFM) module, we conduct experiments using different MFM variants. As shown in Tab.~\ref{table5}, \textit{w/ Mamba} applies the standard Mamba block~\cite{mamba} with forward and backward scanning, \textit{w/ Mamba v2} incorporates an additional region-level mixed-scanning strategy, while \textit{w/ Mamba v3} further integrates a token-level mixed-scanning strategy. \textit{Full Model} denotes the complete quad-scale scanning strategy. The results demonstrate that progressively introducing scanning paths at different levels consistently enhances performance, with the full model achieving the best results. Specifically, QuadFusion outperforms the baseline by \textbf{7.1\%}/\textbf{6.0\%} in PR/SR metrics. Compared to the standard Mamba~\cite{mamba}, QuadFusion achieves notable gains of \textbf{2.0\%}/\textbf{1.5\%} in PR/SR, highlighting the effectiveness of the proposed multiscale fusion approach.
\begin{table}
\centering
\caption{Comparison of different model components.}
\setlength{\tabcolsep}{1mm}
\resizebox{1\linewidth}{!}{
\renewcommand\arraystretch{1.3}{
\begin{tabular}{c|ccc|c|c|c|c}
    \toprule
    \multirow{2}{*}{\textbf{Variant}} & \multicolumn{3}{c|}{\textbf{Scanning-level}} &  \multirow{2}{*}{\textbf{PR}} & \multirow{2}{*}{\textbf{SR}}  & \multirow{2}{*}{\textbf{Params}} & \multirow{2}{*}{\textbf{FLOPs}}\\
    &\textbf{Modal} & \textbf{Region} & \textbf{Token} &  & & &\\
    \hline\hline
    Baseline& - & - & - & 57.0 & 38.7 & 108.6M & 115.2G  \\
    w/ Mamba& \checkmark & - & -  & 62.1 & 42.9 & +2.2M& +1.5G\\
    w/ Mamba v2  & \checkmark & \checkmark & -  & 63.0 & 43.5& +3.0M & +2.1G\\
    w/ Mamba v3  & \checkmark & - & \checkmark & 63.4 & 43.5 & +3.2M & +2.2G \\
    Full Model(QuadFusion) & \checkmark & \checkmark & \checkmark & 
     64.1&
     44.4&
    +4.2M & +3.3G
    \\
    \bottomrule
\end{tabular}}}

\label{table5}
\end{table}

\noindent\textbf{Verification of QuadFusion generality.} We integrate QuadFusion into various trackers and evaluate their performance. As shown in Tab.~\ref{different_base}, QuadFusion consistently improves the performance of all base trackers, confirming its general applicability.
\begin{table}
\centering
\caption{Performance with and without QuadFusion for different trackers.}
\setlength{\tabcolsep}{12mm}
\resizebox{1\linewidth}{!}{
\renewcommand\arraystretch{1}{
\begin{tabular}{c|c|c}
    \toprule
    \textbf{Base Tracker}&  \textbf{PR} & \textbf{SR} \\
    \hline\hline
    OSTrack~\cite{ye2022joint} &  57.0 & 38.4 \\
    w/ QuadFusion &  64.1 & 44.4 \\
    \hline
    ARTrack~\cite{wei2023autoregressive}&  56.4& 38.5 \\
    w/ QuadFusion&  62.9 & 42.3 \\
    \hline
    ToMP~\cite{tomp}&  55.9 & 37.1 \\
    w/ QuadFusion &  60.2 & 41.4 \\
    \bottomrule
\end{tabular}}}

\label{different_base}
\end{table}

\noindent\textbf{Efficiency analysis}.
We compare QuadFusion with two alternative Transformer-based fusion modules. Transformer-A concatenates RGB with each modality and processes them through three ViT blocks. Transformer-B concatenates all modalities together and feeds them through a single ViT block.
As shown in Tab.~\ref{efficiency}, QuadFusion outperforms both alternatives in terms of PR and SR, while maintaining a smaller computational load, demonstrating its superior efficiency. Additionally, QuadFusion's efficiency is compared with other models in the \textbf{Supplementary Material}.

\begin{table}
\centering
\caption{Performance comparison of QuadFusion with Transformer-based fusion modules.}
\setlength{\tabcolsep}{2mm}
\resizebox{1\linewidth}{!}{
\renewcommand\arraystretch{1}{
\begin{tabular}{c|c|c|c|c|cc|c}
    \toprule
    \multirow{2}{*}{\textbf{Method}}&  \multirow{2}{*}{\textbf{PR}} & \multirow{2}{*}{\textbf{SR}}& \multirow{2}{*}{\textbf{Params}}& \multirow{2}{*}{\textbf{Flops}} & \multicolumn{2}{|c|}{\textbf{GPU.Mem}}&\multirow{2}{*}{\textbf{FPS}} \\
    &&&&& \textbf{Train}&\textbf{Test}&\\
    \hline\hline
    Baseline &  57.0 & 38.7 & 108.6M & 115.2G & 16.4G & 1.3G& 29.1\\
    w/ Transformer-A &  61.7 & 42.7 &  +51.5M & +15.4G& +5.3G &+0.3G&26.3\\
    w/ Transformer-B & 62.4 & 43.0 & +24.1M& +26.4G & +8.2G & +0.5G&24.5\\
    w/ QuadFusion &  64.1 & 44.4 & +4.2M & +3.3G & +2.2G & +0.1G& 28.8\\
    \bottomrule
\end{tabular}}}

\label{efficiency}
\end{table}

\subsection{Visualization Analysis}

\begin{figure}[ht]
\centering
\includegraphics[width=1\columnwidth]{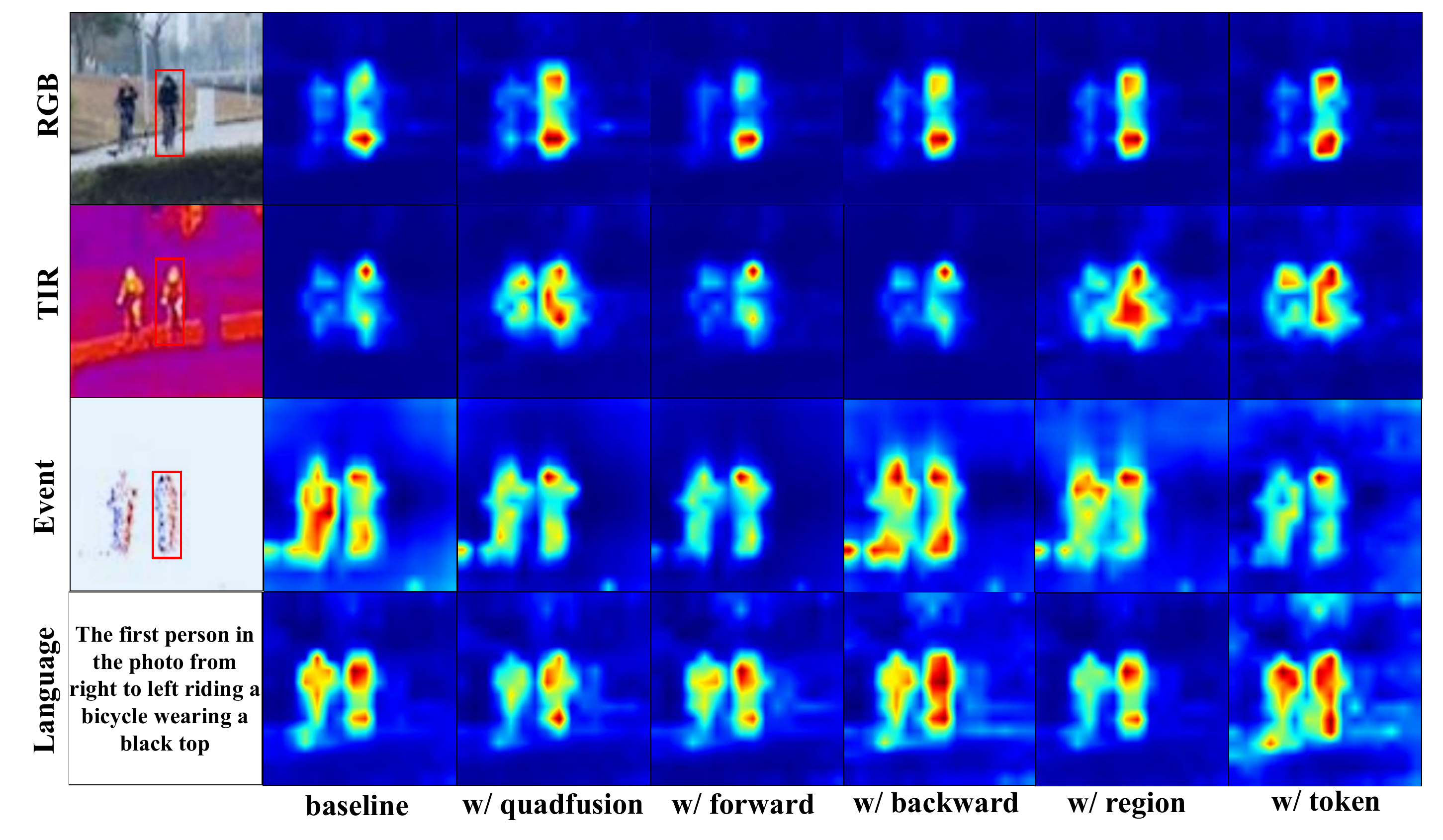} 
\caption{Visualization of different scanning scale.}
\label{vis_fea}
\end{figure}

\noindent\textbf{Visualization of different scanning scales}. 
As shown in Fig.~\ref{vis_fea}, we visualize features from the \textit{right1st\_rider} sequence to analyze the impact of different scanning scales. The baseline model introduces significant noise in both event and language modalities, whereas our multi-scale fusion approach achieves accurate representations across all modalities.
Notably, different fusion scales exhibit distinct behaviors. Due to the inherent sparsity of event data, token-level and region-level scanning introduce residual noise. In the language modality, token-level scanning captures spurious correspondences with irrelevant targets, likely caused by ambiguity arising from linguistic sparsity. In contrast, modal-level scanning integrates sentence-level context, enhancing semantic clarity.
These findings highlight the necessity of multi-scale interaction in multimodal fusion, as different modalities respond differently to interaction granularity. 
Additional visualization examples are provided in the \textbf{Supplementary Material}.

\noindent\textbf{Visualization of tracking results}. We also perform a qualitative analysis of the performance of QuadFusion compared to existing trackers. Visualizations and detailed analysis are provided in \textbf{Supplementary Material}.

\section{Conclusion}
In this work, we introduce QuadTrack600, the first comprehensive multimodal visual tracking benchmark dataset with 600 high-quality video sequences across four modalities: RGB, thermal infrared, event, and language. This dataset contribute to establish a new standard for multimodal visual tracking and provide a foundation for future research, emphasizing the importance of leveraging diverse information sources for robust tracking in complex environments.
We also propose an new approach QuadFusion that designs a multiscale fusion Mamba to enable effective interaction between the four modalities while overcome computational load. Extensive experiments on QuadTrack600 and bi-modal datasets like LasHeR, VisEvent, and TNL2K validate the effectiveness of QuadFusion, demonstrating significant performance improvements.\\
\textbf{Limitation.}The QuadTrack600 is the first quad-modal tracking benchmark that provides a unified platform for existing bimodal tracking tasks. However, we are currently focusing on outdoor scenes, integrating only RGB, TIR, event, and language modalities. In the future, we will consider introducing more modalities, e.g., depth modality, radar, sonar, etc., to build scenario-specific general multimodal tracking.



\section{Supplementary Material}
\subsection{Details of QuadTrack600 Benchmark}
\subsubsection{Evaluation Metrics}
Due to accurate alignment of three modality data, all the trackers are run in one-pass evaluation (OPE) protocol and evaluated by success rate (SR) and precision rate (PR), which are widely used in existing single-modal and bi-modal tracking~\cite{li2021lasher,wang2023visevent,wang2021towards}. 
Overall performance for all sequences and attribute-based performance for attribute-specific sequences are considered.
\begin{itemize}
    \item Success rate (SR). Success rate (SR) measures the ratio of tracked frames, determined by the Interaction-over-Union (IoU) between the predicted bounding box and ground truth is greater than some threshold. With different overlap thresholds, a success plot can be obtained, which can employ the area under the curve to calculate representative SR scores.
    \item Precision rate (PR). PR is to calculate the percentage of frames for which the distance between the predicted location and the ground truth is calculated to be within a threshold $\tau$. $\tau$ is set to 20 in our benchmark.
\end{itemize}

\begin{table*}
\centering
\caption{Attributes defined in QuadTrack600.}
\setlength{\tabcolsep}{2.8mm}{
\resizebox{0.8\textwidth}{!}{
\renewcommand\arraystretch{1.1}{
\begin{tabular}{l|l}
    \toprule
    \textbf{Attributes} & \textbf{Description}  \\
    \hline\hline
    \textbf{01. PO} & Partial Occlusion - the target object is partially occluded  \\
    \textbf{02. TO} &  Total Occlusion - the target object is totally occluded\\
    \textbf{03. HO} & Hyaline Occlusion - the target is occluded by hyaline object \\
    \textbf{04. OV} &  Out-of-View - the target leaves the camera field of view \\
    \textbf{05. VC} &  Viewpoint Change -  changes of viewpoint of the target\\
    \textbf{06. CM} &  Camera Motion - the target object is captured by moving camera\\
    \textbf{07. BC} & Background Clutter - the background information which includes the target is messy \\
    \textbf{08. SA} & Similar Appearance - there are objects of similar appearance near the target\\
    \textbf{09. LI} & Low Illumination - the illumination in the target region is low\\
    \textbf{10. OE} & Over Exposure -  the target object is in an overexposed environment\\
    \textbf{11. IV} & Illumination Variations - Illumination Variations in the background  of the target object\\
    \textbf{12. LR} & Low Resolution - low resolution of target objects in images\\
   \textbf{13. DEF} & Deformation - non-rigid object deformation \\
    \textbf{14. TC} & Thermal Crossover - the target object has the same temperature as its surroundings or other objects\\
    \textbf{15. FL} & Frame Lost - some thermal infrared frames are lost\\
    \textbf{16. FM} & Fast Motion -  the motion of the ground truth between two adjacent frames is large than 20 pixels\\
    \textbf{17. NM} & No Motion -  the target object is in a no motion state\\
    \textbf{18. MB} & Motion Blur - motion of the target object causes blurring of the picture\\
    \textbf{19. SV} & Scale Variation -  the ratio of the first bounding box and the current bounding box is out of the range [0.5,2]\\
    \textbf{20. ARC} & Aspect Ratio Change - the ratio of bounding box aspect is outside the range [0.5,2]\\
    \textbf{21. BOM} & Background Object Motion - influence of background object motion for Event camera \\
    \bottomrule
\end{tabular}}}}

\label{attributes_description}
\end{table*}

\subsubsection{Attribute Definition} 
To comprehensively evaluate the performance of the tracker under various challenges, we also define a total of 21 challenge attributes. These attributes cover several key aspects: firstly, occlusion-related challenges, including partial occlusion (PO) and total occlusion (TO); secondly, illumination-related challenges, such as low illumination (LI) and over-exposure (OE); thirdly, challenges of modality limitations, such as thermal crossover (TC) and background object motion (BOM); and fourthly, challenges associated with the target itself, such as fast movement (FM) and scale variation (SV). Other challenge attributes can be found in Tab.~\ref{attributes_description}.

\subsection{Experiments on Bi-modal Tracking Task}
QuadTrack600 dataset holds great potential for advancing generalized multimodal tracking. It not only provides a quad-modal tracking benchmark but also includes three distinct bi-modal tracking benchmarks, including QuadTrack600-RT, QuadTrack600-RE, and QuadTrack600-RL representing RGB-T, RGB-E, and RGB-L tracking, respectively. These subsets can provide a novel dimension of comparison to the existing field of bi-modal tracking due to the consistency of the scenes. Similarly, we introduce a large number of state-of-the-art bimodal trackers to evaluate each of the three subsets to validate the challenges of the subsets. In addition, we evaluate the proposed QuadFusion in three bi-modal tracking subsets and three public bi-modal tracker datasets, including LasHeR (RGB-T), VisEvent (RGB-E), and TNL2K (RGB-L), to demonstrate the adaptability of the general tracker for bi-modal scenes.
\subsubsection{Evaluation on  RGB-T Subset}
\begin{table}
\centering
\caption{Performance of RGB-T tracker on QuadTrack600-RT subset.}
\setlength{\tabcolsep}{1mm}
\resizebox{0.48\textwidth}{!}{
\renewcommand\arraystretch{1.3}{
\begin{tabular}{c |c |cc |ccc}
    \toprule
    \multirow{2}{*}{\textbf{Method}} &  \multirow{2}{*}{\textbf{Pub. Info.}} & \multicolumn{2}{c|}{\textbf{QuadTrack600-RT}}  & \multicolumn{3}{c}{\textbf{LasHeR}}\\
    & & \textbf{PR} & \textbf{SR} & \textbf{PR} & \textbf{NPR} & \textbf{SR}\\
    \hline\hline
       
       DMCNet~\cite{lu2022duality} & TNNLS-2022 & 46.4 & 29.4 & 49.0 & 43.1 & 35.5\\
       HMFT~\cite{Zhang_CVPR22_VTUAV} & CVPR-2022 & 44.7 & 28.0 & - & - & - \\
       ViPT~\cite{zhu2023visual} & CVPR-2023 & 51.2 & 35.6 & 65.1& - & 52.5\\
       TBSI~\cite{hui2023bridging}  & CVPR-2023 & 59.2 & 40.0 & 69.2 & 65.7 & 55.6\\    
       CAT++~\cite{liu2024rgbt} & TIP-2024 & 45.8 & 30.4 & 50.9 & 44.4 & 35.6\\
       Un-Track~\cite{wu2024single} & CVPR-2024 & 45.1 & 32.2 & 66.7& - & 53.6\\
       BAT~\cite{cao2024bi} & AAAI-2024 & 61.5 & 42.9 & 70.2 & 66.4 & 56.3\\
       SDSTrack~\cite{hou2024sdstrack} & CVPR-2024 & 44.9 & 30.1 & 66.5 & - & 53.1 \\
       \cellcolor{lightgray}QuadFusion(Ours) & \cellcolor{lightgray}- & \cellcolor{lightgray}62.6& \cellcolor{lightgray}43.1& \cellcolor{lightgray}70.9& \cellcolor{lightgray} 66.8& \cellcolor{lightgray}56.6\\
    \bottomrule
\end{tabular}}}

\label{table_rt}
\end{table}


As shown in Tab.~\ref{table_rt}, we select 8 trackers for testing on QuadTrack600-RT. We can see that the performance of mdnet and dimp based trackers is low, the performance of transformer based trackers is relatively better, BAT and TBSI achieve high performance compared to general purpose trackers, due to the specially designed modules. Our method achieves good performance on both the QuadTrack600-RT as well as LasHeR datasets, validating the adaptability of our method to RGB-T tracking scenarios.

\subsubsection{Evaluation on RGB-E Subset}
\begin{table}
\centering
\caption{Performance of RGB-E tracker on QuadTrack600-RE subset. }
\setlength{\tabcolsep}{1mm}
\resizebox{0.48\textwidth}{!}{
\renewcommand\arraystretch{1.3}{
\begin{tabular}{c | c | cc | cc}
    \toprule
    \multirow{2}{*}{\textbf{Method}} &  \multirow{2}{*}{\textbf{Pub. Info. }} & \multicolumn{2}{c|}{\textbf{QuadTrack600-RE}} & \multicolumn{2}{c}{\textbf{VisEvent}} \\
    & & \textbf{PR} & \textbf{SR} & \textbf{PR} & \textbf{SR}\\
    \hline\hline
     ViPT~\cite{zhu2023visual} & CVPR-2023 & 49.9 & 33.9 & 75.8 & 59.2 \\ 
     
     COHA~\cite{zhu2023cross} & ICCV-2023 & 49.3 & 30.5 & - & -\\
     AFNet~\cite{zhang2023frame} & CVPR-2023 & 47.3 & 31.2 & 59.3 & 44.5 \\
     Un-Track~\cite{wu2024single} & CVPR-2024 & 51.1 & 35.1 & 76.3  & 59.7\\
     SDSTrack~\cite{hou2024sdstrack} & CVPR-2024 & 52.2 & 36.2 & - & -\\
    \cellcolor{lightgray} QuadFusion(Ours) & \cellcolor{lightgray}- & \cellcolor{lightgray} 61.7& \cellcolor{lightgray} 42.4& \cellcolor{lightgray} 76.6& \cellcolor{lightgray}60.2\\
    \bottomrule
\end{tabular}}}

\label{table_re}
\end{table}

As shown in Tab.~\ref{table_re}, we select 5 trackers for testing on QuadTrack600-RE. Benefit from our multiscale scanning strategy, our method  enhance event modality with sparser information quantity, and our method has the best performance of 61.7\%/42.4\%. In addition our method achieves good results on the VisEvent dataset, which fully demonstrates the applicability of our method to RGB-E tracking scenarios.

\subsubsection{Evaluation on RGB-L Subset}
\begin{table}
\centering
\caption{Performance of RGB-L tracker on QuadTrack600-RL subset. }
\setlength{\tabcolsep}{1mm}
\resizebox{0.48\textwidth}{!}{
\renewcommand\arraystretch{1.3}{
\begin{tabular}{c | c | cc | cc}
    \toprule
    \multirow{2}{*}{\textbf{Method}} &  \multirow{2}{*}{\textbf{Pub. Info.}} & \multicolumn{2}{c|}{\textbf{QuadTrack600-RL}} & \multicolumn{2}{c}{\textbf{TNL2K}} \\
    & & \textbf{PR} & \textbf{SR} & \textbf{P} & \textbf{AUC}\\
    \hline\hline
     VLT$_{TT}$~\cite{guo2022divert} & NeurIPS-2022 & 42.8 & 
     30.2 & 53.3 & 53.1\\
     JointNLT~\cite{zhou2023joint} & CVPR-2023 & 40.1 & 25.8 & 58.1 & 56.9\\
     CiteTracker~\cite{li2023citetracker} & ICCV-2023 &  50.7& 34.1 &- & - \\
     MMTrack~\cite{zheng2023towards} & TCSVT-2023 & 54.4 & 35.9 & 59.4 & 58.6\\
     All-in-One~\cite{zhang2023all} & ACM MM-2023 & 44.0 & 28.4 & 57.2 & 55.3\\
     \cellcolor{lightgray}QuadFusion(Ours) & \cellcolor{lightgray}- & \cellcolor{lightgray} 59.3& \cellcolor{lightgray}40.8& \cellcolor{lightgray} 61.1&\cellcolor{lightgray} 59.4\\
    \bottomrule
\end{tabular}}}
\label{table_rl}
\end{table}

As shown in Tab.~\ref{table_rl}, we select 5 trackers for testing on QuadTrack600-RL. We can see that the overall performance of these trackers is poor, which suggests that the RL subset is challenging, and benefit from our multiscale fusion module, our method achieves better performance on the RL subset compared to the other trackers. In addition, our method also achieves relatively good results on TNL2K, which demonstrates the adaptability of our tracker to bimodal scenarios.

\begin{figure*}[ht]                                                                    
\centering
\includegraphics[width=2\columnwidth]{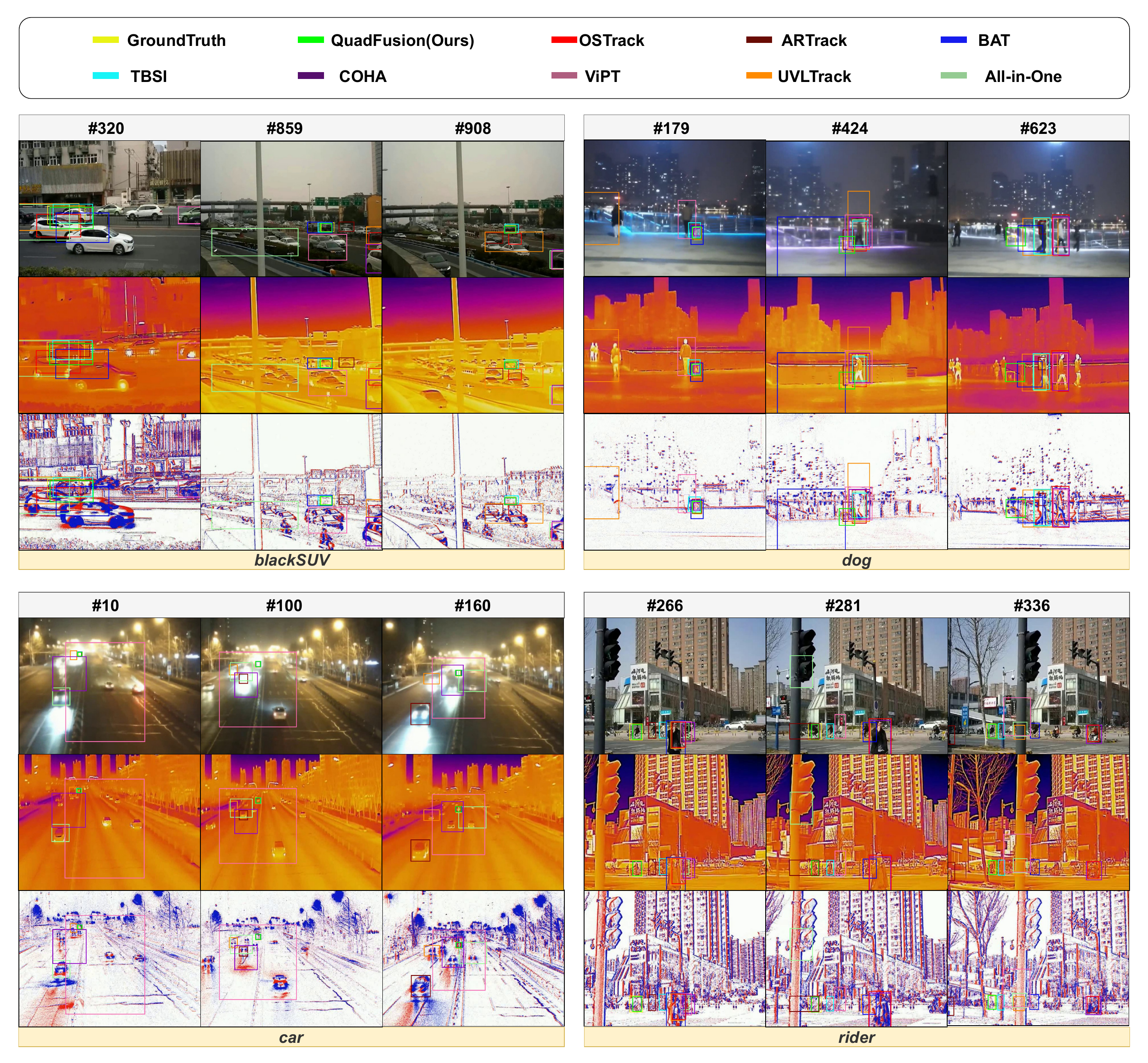} 
\caption{Qualitative comparison of QuadFusion and other methods on the QuadTrack600.}
\label{vistrack}
\end{figure*}

\subsection{Efficiency Analysis}
As shown in Tab.~\ref{efficiency1}, we compare QuadFusion with several other attention-based fusion methods (all methods use quad-modal data as inputs). While QuadFusion achieves competitive performance with reasonable computational overhead, it underperforms in terms of inference speed, this is because Mamba has not yet adapted to existing acceleration hardware.
\begin{table}
\centering
\caption{Comparison of QuadFusion and other methods on QuadTrack600.}
\setlength{\tabcolsep}{3mm}
\resizebox{1\linewidth}{!}{
\renewcommand\arraystretch{1}{
\begin{tabular}{c|c|c|c|c|c}
    \toprule
    \textbf{Method}&  \textbf{PR} & \textbf{SR}& \textbf{Params}& \textbf{Flops} & \textbf{FPS} \\
    \hline\hline
    QuadFusion &  64.1 & 44.4 & 112.8M & 118.5G &  28.8\\
    TBSI &  61.3 & 62.4 & 297.8M & 130.5G & 27.0\\
    COHA &  50.9 & 35.0 & 230.2M & 135.7G & 25.8\\
    All-in-One & 46.3 & 29.8 & 201.5M & 122.6G & 24.2\\
    \bottomrule
\end{tabular}}}

\label{efficiency1}
\end{table}

\subsection{Visualization of Different Scanning Scales}
\begin{figure}[ht] 
\centering  
\includegraphics[width=1\columnwidth]{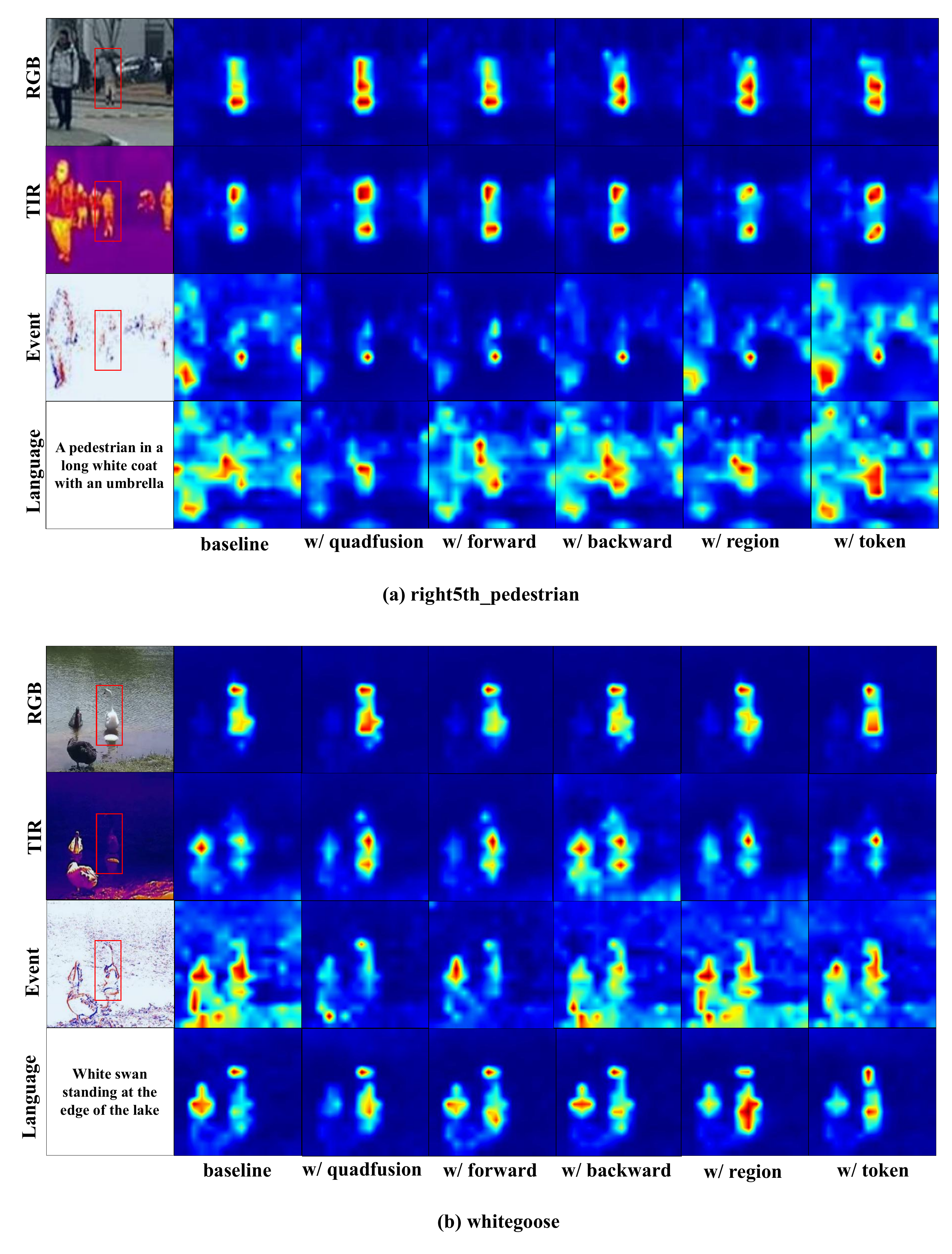} 
\caption{Visualization of different scanning scales.}
\label{vis_fea1}
\end{figure}

As illustrated in Fig.~\ref{vis_fea1}, we perform feature visualization of several representative sequences to further validate the effectiveness of Quadfusion and different scanning scales. The following annotations describe specific configurations: \textit{baseline} indicates features extracted without the multiscale fusion Mamba module, \textit{w/ quadfusion} refers to features derived using the multiscale fusion Mamba module, \textit{w/ forward} denotes features obtained from modal-level forward scanning, \textit{w/ backward} represents features derived from modal-level backward scanning, \textit{w/ region} indicates features generated through region-level scanning and \textit{w/ token} corresponds to features from token-level scanning.

It can be observed that \textit{w/ quadfusion} significantly enhances the object representation in all four modalities while effectively suppressing the noise compared to its absence, while effectively suppressing noise. For example, as shown in the second and third columns in sequence (a), the RGB and TIR modalities are more responsive to the object's features after the multiscale fusion Mamba module, while clearly suppressing the background noise in the event and language modalities.

Different modalities exhibit varying responses to the object at different scanning scales. In sequence (a), both the RGB and TIR modalities show higher information density, with their responses becoming more significant across the scanning scales, from the fourth to the seventh columns.  
In contrast, the event modality has lower information density, with weaker responses in the sixth and seventh columns during fine-scale region-level and token-level scanning. However, during coarse-scale modal-level scanning (second, fourth and fifth columns), the response of the event modality becomes more significant.
In sequence (b), the language modality provides a fine-grained description of the object. While its response is not significantly enhanced during coarse-scale modal-level scanning (second, fourth and fifth columns), it becomes more significant during fine-scale region-level and token-level scanning (second, sixth and seventh columns), where it not only strengthens its response to the object but also suppresses background noise.

These experiments demonstrate that due to the differences in information density across modalities, their discriminative power varies at different scanning scales. This highlights the necessity of multi-scale scanning to complement the information gaps between individual modalities.


\subsection{Visualization of Tracking Results}
As shown in Fig.~\ref{vistrack}, we perform a qualitative comparison between our tracker and eight other trackers. We selecte four representative sequences from the QuadTrack600 dataset, including various challenges such Partial Occlusion, Viewpoint Change , Camera Moving, Background Clutter, Similar Appearance, Frame Lost, Fast Motion, Background Object Motion, etc. , to compare the performance of different approaches. For example, in the \textbf{\textit{car}} sequence, car traveling in overexposed environments, the RGB modality is almost ineffective, our method fully utilizes the Event modality to fully complement the acquisition of information to obtain the trajectory of the car. In the \textbf{\textit{rider}} sequence, the background is more cluttered and there is also occlusion, and the total tracking object is in a group of similar looking objects, and our method obtains more robust tracking by fusing the descriptive information of the tracking object.

\bibliographystyle{IEEEtran}
\bibliography{QuadFusion}

\end{document}